# A Targeted Acceleration and Compression Framework for Low-bit Neural Networks

Biao Qian, Yang Wang

Hefei University of Technology, China

yangwang@hfut.edu.cn

1. Abstract

   1-bit deep neural networks (DNNs), of which both the activations and weights are binarized, are attracting more and more attention due to their high computational efficiency and low memory requirement. However, the drawback of large accuracy dropping also restricts its application. In this paper, we propose a novel Targeted Acceleration and Compression (TAC) framework to improve the performance of 1-bit deep neural networks. We consider that the acceleration and compression effects of binarizing fully connected layers are not sufficient to compensate for the accuracy loss caused by it. In the proposed framework, the convolutional and fully connected layer are separated and optimized individually. For the convolutional layers, both the activations and weights are binarized. For the fully connected layers, the binarization operation is replaced by network pruning and low-bit quantization. The proposed framework is implemented on the CIFAR-10, CIFAR-100 and ImageNet (ILSVRC-12) datasets, and experimental results show that the proposed TAC can significantly improve the accuracy of 1-bit deep neural networks and outperforms the state-of-the-art by more than 6 percentage points.

2. Introduction

   Recently, deep neural networks (DNNs) have widely applied in computer vision tasks, such as image classification [17] [18], object detection [24] and visual recognition [25][26][27][28][29][30][31][32][33] so on. Under the circumstances, the application of deep neural networks on mobile devices is gaining more and more attention. However, great results of DNNs are usually accompanied by large models and high computational complexity, which limits their application, due to the limited memory and computation resources of the mobile devices. For the mobile devices, small and fast neural networks are more helpful on real-world application. Therefore, it's increasingly becoming crucial to accelerate and compress deep neural networks.

   In order to effectively accelerate and compress deep neural networks, many works have been proposed, such as pruning [1] [2] [3], low-bit quantization [4] [8] [9] [13] [16], extremely binarization [11] [12] and so on. Among them, 1-bit deep neural networks, which work for training DNNs with binary weights or even activations, are receiving more and more attention. In BinaryConnect [10], the weights are binarized to either +1 or -1 to save memory, and multiplications are turned into additions and subtractions to speed up the computation. In [12], binary weighted networks (BWN) proposed adds a scaling factor to the binary weights to improve network performance. Besides, BNN [11] and XNOR-net [12] further binarize both the activations and weights to speed up network computation. In this case, the floating-point multiply-accumulate operations can be replaced by the XNOR and bit-count operations, which greatly saves computational resource. However, 1-bit deep neural networks tolerate the sharp reduction of prediction accuracy when both activations and weights are binarized. For instance, on the ImageNet dataset with AlexNet, XNOR-net's prediction accuracy dramatically drops from 56.6% to 44.2% [12].

For the traditional deep neural networks, most binarization methods do not take the differences between convolutional layer and fully connected layer into account, or just consider fully connected layer as a special convolutional layer. As we know, the convolutional layer is often computationally intensive, while the fully connected layer has most of network parameters. For example, as shown in table 1, for AlexNet, the convolutional layers occupy more than 90% of the overall computational resources, while the fully connected layers hold more than 95% of the total parameters. Therefore, the convolutional layers tend to be accelerated, while fully connected layers tend to be compressed.

Table 1: Memory usage and FLOPs of AlexNet

| Layer | Weights | FLOPs |
|-------|---------|-------|
| conv1 | 35K     | 211M  |
| conv2 | 307K    | 448M  |
| conv3 | 885K    | 299M  |
| conv4 | 663K    | 224M  |
| conv5 | 442K    | 150M  |
| fc6   | 38M     | 75M   |
| fc7   | 17M     | 34M   |
| fc8   | 4M      | 8M    |
| Total | 61M     | 1.5G  |

Based on the above facts, current binarization methods encounter at least three limitations. Firstly, binarizing the whole network usually causes severe accuracy degradation. Secondly, the impact of binarizing fully connected layers on the acceleration of the entire network is much smaller than that of binarizing convolutional layers, which indicates that the binarization of fully connected layers is not very necessary. Lastly, binarizing fully connected layers except the last layer can results in the degradation of the whole compression rate. In contrast, our results show that accelerating and compressing fully connected layers using other compression methods instead of binarization can obtain more significant consequents.

In this paper, to overcome the aforementioned issues, we propose a novel framework named Targeted Acceleration and Compression (TAC), which separates the fully connected layers from the convolutional layers and optimizes them individually. In this work, our main goal is to improve the accuracy of 1-bit deep neural networks. To achieve that, the implement of the TAC can be regarded as two steps: accelerating the convolutional layer and compressing the fully connected layer. In the first step, the binarization method is adopted to accelerate convolutional layers as previous works do, where the real-value multiply-accumulate operations in convolutional computation are converted into the XNOR and bit-count operations to reduce the computational complexity. In the second steps, we use network pruning and low-bit quantization instead of binarization to compress fully connected layers to reduce the redundancy on network structure and data storage, respectively. Our framework is shown as Figure 1.

To the best of our knowledge, this paper is the first attempt to separately consider convolutional layers and fully connected layers of deep neural networks for targeted acceleration and compression. Major contributions of this paper are as follows: we introduce a novel TAC framework for targeted acceleration and compression of deep neural networks, where neural networks are optimized separately according to different network structures. As a result, our framework achieves a great trade-off among computational complexity, compression rate and network accuracy.

## 3. Related Work

In this section, we mainly review the methods related to pruning and quantization.

**Network pruning:** Network pruning measures the redundancy on network structures like connections, neurons and filters with different rules, and removes unimportant parts. In [1], all connections with weights below a certain threshold are removed from the pre-trained network. [2] imposes L1 regularization on the scaling factors in batch normalization (BN) layers, then prunes those channels with small scaling factors. [3] uses L1-norm to select unimportant filters and physically prune them. [5] proposes a global, dynamic pruning method that adds more flexibility to pruning of the network.

**Quantization:** The goal of quantization is to reduce the redundancy on storage of network parameters. [4] performs trained quantization of network parameters using k-means clustering and parameter sharing without the performance loss. [6] proposes entropy-constrained network quantization schemes, further improving network compression ratio. [7] proposes incremental network quantization, which includes three operations: weight partition, group-wise quantization and re-training. These three operations are repeated in an iterative manner until all the weights are converted into either power of two or zero. Besides, low-bit quantization also becomes popular. [8] [9] quantize real-valued weights into ternary weights, where weights are restricted to {+1, 0, -1}. [16] proposes a Two-Step Quantization framework for low-bit quantized neural networks, whose motivation is to decouple the weight quantization from activation quantization.

**Binarization:** Binarization means that weights or activations of neural networks are restricted to two possible values. Weight binarization can greatly reduce model size. [10] proposes BinaryConnect to learn deep neural networks with binary weights, however, their method works well only on small datasets, such as MNIST, CIFAR10 and SVHN. Binary-Weight-Networks (BWN) proposed in [12] achieves comparable accuracy with the full precision counterpart on ImageNet dataset. In the work of [14], they reveal the close connection between inner-product preserving hashing and binary weight neural networks. Based on this view, training binary weight networks can be transformed into a hashing problem.

Besides, activation binarization can effectively reduce computational complexity, where the network computation simply contains the XNOR and bit-count operations. Binarized Neural Networks (BNN) proposed in [11] adopts sign function to restrict both the activations and weights to either +1 or -1, achieving comparable accuracy on small dataset like CIFAR-10. In the work of [12], the XNOR-net proposed introduces a scaling factor to improve the performance of binary neural networks. On large dataset like ImageNet, the XNOR-net achieves much higher accuracy than BNN, however, still tolerates a large drop compared with the full precision counterpart. In order to further improve the accuracy of binary neural networks, [15] proposes a novel scheme for binarizing CNNs. They approximate full-precision weights with the linear combination of multiple binary weight bases and introduce multiple binary activations, achieving prediction accuracy comparable to its full-precision counterpart on ImageNet dataset.

## 4. Method

In the section, we introduce our Targeted Acceleration and Compression (TAC) framework in detail.

Considering the binarization of fully connected layers is not very necessary and can cause the accuracy loss, our TAC optimizes convolutional layers and fully connected layers individually and regards the process as two steps: accelerating convolutional layers and compressing fully connected layers. Figure 1 illustrates our TAC framework. First, the initial network is trained with binary activations and weights of convolutional layers. Next, network pruning and low-bit quantization are adopted successively to replace the binarization operation to compress fully connected layers based the above trained network. Finally, an accelerated and compressed network is obtained. More details are as follows.

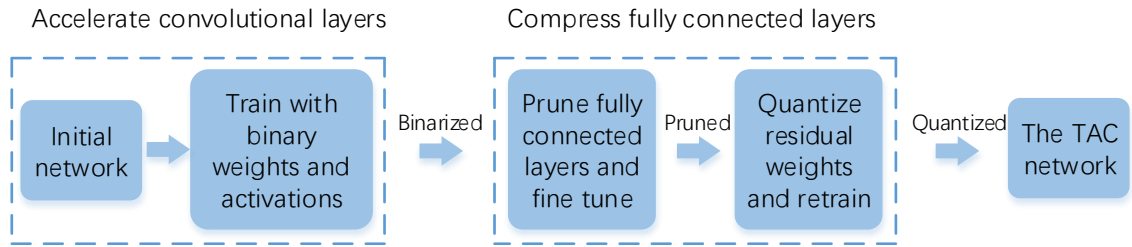

Figure 1: Illustration of the proposed targeted acceleration and compression framework

**Accelerating:** Binarization can greatly speed up network computation and reduce model size. In this work, we apply binarization methods [11] [12] to accelerate convolutional layers, where both the activations and filters of each convolutional layer are restricted to two possible values (e.g. +1 or -1). Then convolutional computations can significantly be accelerated by converting the real-value multiply-accumulate operations into the XNOR and bit-count operations [11], as shown in Figure 2. Specifically, as [11] [12] do, the binarization operation can be implemented through a sign function:

$$x^b = sign(x) = \begin{cases} +1 & if\ x \geq 0 \\ -1 & otherwise \end{cases}$$

where $x^b$ is the binarized value, and $x$ is the full-precision value. The sign function is simple to implement and works well.

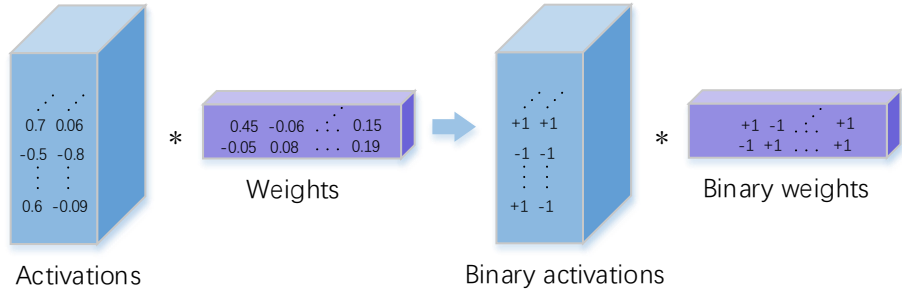

Figure 2: The binarization process of convolutional layers. Blue and purple boxes represent activations and filters, respectively.

**Compressing:** Network pruning and low-bit quantization have been widely used to compress DNNs. In this work, network pruning [1] is utilized to sparse fully connected layers based on the above trained binary network. As shown in [1], fully connected layers are less sensitive to the pruning than convolutional layers and the unimportant connections can be removed from the network directly instead of becoming zeros. For the remaining weights, low-bit quantization is applied to further compress the pruned network by using as few bits as possible to represent each weight, where the weights are replaced by the index of the corresponding quantization levels, as proposed in [4].

Figure 3 illustrates the pruning and low-bit quantization operations using a 4×4 fully connected layer. We set the pruning rate and the bit width as 0.25 and 2, respectively. The pruning is first applied to remove the four connections with weights below a threshold from the network. We then partition the unpruned weights using k-means clustering, and set the clustering centroids as the quantization levels. Finally, we replace all the weights of each partition using the index of the corresponding level values.

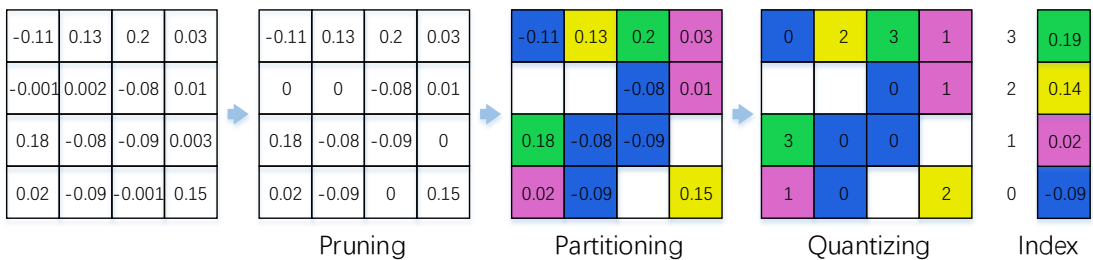

Figure 3: The pruning and quantization operations of fully connected layers. Different colors indicate the corresponding partition.

## 5. Experiments

We performed extensive experiments on CIFAR-10, CIFAR-100 and ImageNet datasets with VGG-9 and AlexNet architectures.

### 5.1 Datasets and Implement Details

In this section, we briefly introduce the datasets, network structures and experiment settings in our experiments.

**CIFAR-10/100 [22]:** This dataset consists of a training set of 50,000 and a test set of 10,000 color images with resolution 32 × 32. CIFAR-10 and CIFAR-100 contain 10 and 100 classes images, respectively. The network architecture that we adopt is a variant of VGG-net [18], denoted as VGG-9 [10], which consists of 6 convolutional layers and 3 fully connected layers. The network is trained with ADAM [21] with momentum 0.9 and weight decay 0.00001. The learning rate starts at 0.001, and is divided by 2 every 60 epochs. During the fine-tuning, the learning rate is set as 1e-8. The network is trained for 400 epochs with batch size equals to 128.

**ImageNet (ILSVRC12) [23]:** This dataset is a large image classification dataset, which contains about 1.2M training images and 50K validation images of 1,000 classes. The classification performance is evaluated by using Top-1 and Top-5 accuracies. For the ImageNet dataset, training images are rescaled to the size of 256×256, and a 224×224 (227×227 for AlexNet) crop was randomly sampled from an image or its horizontal flip. We adopt a variant of AlexNet [17] architecture with batch normalization, which consists of 5 convolutional layers and 3 fully connected layers. The network is trained with ADAM [21] with momentum 0.9 and weight decay 0.00001. The learning rate starts at 0.001, and decayed by 10 every 25 epochs, and is set as 1e-8 during the fine-tuning. We trained the network for 100 epochs with batch size equals to 256.

Under the TAC framework, we adopt the BNN and XNOR-net methods proposed in [11] [12] to binarize both the activations and weights of convolutional layers, and network pruning and quantization in [1] [4] to compress fully connected layers. For the hyper-parameters, we set the pruning rate at iterative steps as {0.2, 0.4, 0.6, 0.7, 0.75} and the bit width of quantization as 4. For simplicity, we denote the networks trained with the TAC framework as TAC-BNN and TAC-XNOR according to the binarization methods adopted, respectively.

### 5.2 Experimental Results

To evaluate our proposed method, we compare our TAC with two methods: BNN and XNOR-net. These two networks are representative methods for training 1-bit deep neural networks and achieve the state-of-the-art results.

**VGG-9 on CIFAR-10 and CIFAR-100:** Table 2 shows the classification accuracy on CIFAR-10 and CIFAR-100 datasets. The TAC-BNN and TAC-XNOR improve the accuracy of BNN and XNOR-net significantly. On CIFAR-10 dataset, the TAC-BNN achieves more than 4% accuracy improvement. And both the TAC-BNN and TAC-XNOR improve the accuracy by over 2% on CIFAR-100. Moreover, the TAC achieves a compression rate comparable to or even higher than the BNN and XNOR-net, as shown in Table 3.

Table 2: Accuracy (%) on CIFAR-10 and CIAFR-100

| Model | CIFAR-10 | CIFAR-100 |
|---|---|---|
| Full-Precision | 93.23 | 70.77 |
| BNN | 88.13 | 66.93 |
| TAC-BNN | 92.52 | 69.01 |
| XNOR-net | 91.24 | 66.84 |
| TAC-XNOR | 92.08 | 69.57 |

Table 3: Model size and compression rate on CIFAR-10 and CIAFR-100

| Model | CIFAR-10 | CIFAR-100 |
|---|---|---|

|  | Model Size | Compression rate | Model Size | Compression rate |
|---|---|---|---|---|
| Full-Precision | 53.5M | 1× | 53.8M | 1× |
| XNOR-net/BNN | 1.683M | 31.7× | 1.771M | 30.4× |
| TAC | 1.678M | 31.8× | 1.722M | 31.2× |

**AlexNet on ImageNet:** Table 4 presents the results on the ImageNet dataset with AlexNet. The results show that the TAC-BNN and TAC-XNOR outperform the BNN and XNOR-net significantly. Compared with the BNN and XNOR-net, the TAC improves the Top-1 accuracy by 4.2% and 6.3%, and increases the compression rate from 10.3× to 26.1×.

Table 4: Accuracy (%) on ImageNet

| Model | Top-1 | Top-5 | Model size | Compression rate |
|---|---|---|---|---|
| Full-precision | 57.22 | 80.27 | 232M | 1× |
| Network pruning [1] | 57.23 | 80.33 | 26M | 9× |
| Deep compression [4] | 57.22 | 80.30 | 8.6M | 27× |
| BNN | 37.28 | 62.18 | 22.6M | 10.3× |
| TAC-BNN | 41.49 | 65.02 | 8.9M | 26.1× |
| XNOR-net | 44.20 | 69.20 | 22.6M | 10.3× |
| TAC-XNOR | 50.51 | 74.56 | 8.9M | 26.1× |

**5.3 Efficiency Analysis**

In this section, we analyze the computation efficiency of the TAC by comparing with the XNOR-net [12], network pruning [1] and deep compression [4]. For comparison, we use FLOPs to measure the computational complexity. Following the calculation methods in [20], the FLOPs of the binary layers with the XNOR and bit-counting operations is calculated as 1/64 of the amount of 1-bit multiplication.

Table 5 presents the FLOPs of different methods on AlexNet. Network pruning [1] adopts pruning to learn efficient neural networks, and deep compression in [4] adopts pruning and trained quantization to compress the networks. The XNOR-net binarizes both the activations and weights of the network except the first and last layer. The results show that the TAC-XNOR achieves computational reduction of about 5.5 times. For the whole network, our method improves the accuracy and compression rate with a small increase of FLOPs compared with the XNOR-net. Compared with Network pruning [1] and Deep compression [4], binarizing the convolutional layers also brings significant computational acceleration.

Table 5: Comparison of FLOPs

| Model | FLOPs | Computation saving |
|---|---|---|
| Full-precision | $1.45 \times 10^9$ | 1× |
| Network pruning [1] | $4.35 \times 10^8$ | 3.3× |
| Deep compression [4] | $4.35 \times 10^8$ | 3.3× |
| XNOR-net | $2.38 \times 10^8$ | 6.1× |
| TAC-XNOR | $2.63 \times 10^8$ | 5.5× |

## 6. Conclusion

In this paper, we have proposed a novel framework named Targeted Acceleration and Compression (TAC), where the convolutional and fully connected layer are separated and learned individually. For the convolutional layers, both the activations and weights are binarized to speed up the computations. For the fully connected layers, network pruning and low-bit quantization are adopted to further reduce memory. The final results demonstrate the effectiveness of the proposed TAC and show that the accuracy of the TAC outperforms the state-of-the-art. Our framework improves the performance of 1-bit deep neural networks and facilitates the applications of DNNs on the mobile devices.